\documentclass{article}
\usepackage{ijcai18}

\usepackage{times}
\usepackage{xcolor}
\usepackage{soul}
\usepackage[utf8]{inputenc}
\usepackage[small]{caption}

\usepackage{graphicx}
\usepackage{comment}
\usepackage{subfigure}
\usepackage{multirow}

\usepackage{algorithm}
\usepackage{algorithmicx}
\usepackage{algpseudocode}
\usepackage{amssymb}
\usepackage{amsmath}
\usepackage{epstopdf}
\DeclareMathOperator*{\argmax}{argmax}

\title{On Improving Deep Reinforcement Learning for POMDPs}

\author{
Pengfei Zhu$^1$, 
Xin Li$^2$, 
Pascal Poupart$^3$, 
Guanghui Miao$^4$, 
\\ 
$^1$$^,$$^2$$^,$$^4$ Beijing Institute of Technology, Beijing, China \\
$^3$ Waterloo, Ontario  \\
zhu\_pengfei0408@163.com,
xinli@bit.edu.cn,
ppoupart@uwaterloo.ca,
cortwi@163.com
}

\begin{document}

\maketitle

\begin{abstract}
  Deep Reinforcement Learning (RL) recently emerged as one of the most competitive approaches for learning in sequential decision making problems with fully observable environments, e.g., computer Go. However, very little work has been done in deep RL to handle partially observable environments. We propose a new architecture called Action-specific Deep Recurrent Q-Network (ADRQN\footnote{ADRQN is implemented in Caffe. The source is available at https://github.com/bit1029public/ADRQN.}) to enhance learning performance in partially observable domains. Actions are encoded by a fully connected layer and coupled with a convolutional observation to form an action-observation pair. The time series of action-observation pairs are then integrated by an LSTM layer that learns latent states based on which a fully connected layer computes Q-values as in conventional Deep Q-Networks (DQNs). We demonstrate the effectiveness of our new architecture in several partially observable domains, including flickering Atari games.
\end{abstract}

\section{Introduction}


Deep reinforcement ~\cite{silver2016mastering} learning combines reinforcement learning and deep learning, and has shown great success in solving a number of very challenging tasks that can be easily modeled as conventional reinforcement learning problems, but cannot be solved by conventional reinforcement learning approaches due to the high dimensionality of the state space, e.g., continuous control~\cite{DBLP:journals/corr/LillicrapHPHETS15}, high-dimensional robot control~\cite{DBLP:journals/corr/StadieLA15}, and Atari Learning Environment benchmarks (ALE)~\cite{DBLP:journals/jair/BellemareNVB13}.


Specifically, Deep Q-Networks (DQNs) can be used to effectively and efficiently play many Atari 2600 games~\cite{mnih2015human}. The idea is to extract convolved features of the frames as states and to approximate the Q-function over the states with a deep neural network. For instance, a DQN is trained to estimate the expected value of a policy based on the the past four frames as input in Atari 2600 games. For most Atari 2600 games, four frames are sufficient to approximate the current state of the game. Generally speaking, the Atari games are treated as MDP problems, and DQNs are used to approximate the value functions of these MDPs. However, for some other real-world tasks including some Atari 2600 games, the problem is really a partially observable Markov decision process (POMDP) where the state of the environment may be partially observable or even unobservable, to the point where arbitrarily long histories of observations are needed to extract sufficient features for optimal action selection. Unfortunately, DQN is not suitable for those problems due to the assumption of complete observability of the state.

While POMDPs can naturally model planning tasks with uncertain action effects and partial state observability, finding an optimal policy is notoriously difficult. Some previous POMDP techniques focus on identifying a finite subset of beliefs that are sufficient to approximate all reachable beliefs in order to decrease computational complexity~\cite{DBLP:conf/rss/KurniawatiHL08,DBLP:conf/icml/LiCLW07}. 
\begin{table*}
    \begin{tabular}{|c|c|c|c|}
        \hline
            Model & Input & Problem Addressed & Description \\
            \hline
            DQN & $s_t $ & model-free MDP &
            full knowledge of the state required \\
            \hline
            DBQN & $b_t$ & model-based POMDP
            & updated belief state required \\
            \hline
            DRQN & $\langle o_1,o_2,\cdots,o_t \rangle$ & model-free POMDP & observations as the input\\
            \hline
            \multirow{2}{*}{ DDRQN } &
            $\langle a_0,a_0,\cdots,a_{t-1} \rangle$ &
            \multirow{2}{*}{ model-free POMDP }&
            \multirow{2}{*}{decoupled actions and observations required } \\
             & $\langle o_1,o_2,\cdots,o_t \rangle$ & & \\
            \hline
            ADRQN& $\langle (a_0,o_1),(a_1,o_2),\cdots,(a_{t-1},o_t) \rangle$ & model-free POMDP& action-observation pairs required as input  \\
        \hline
    \end{tabular}
    \caption{A comparison among state-of-the-art deep Q-learning approaches}
    \label{table:modelscomp}
\end{table*}
Recent advances in deep learning suggest a new way of thinking for solving POMDP problems. However, very little work leverages deep reinforcement learning in partially observable environments. Among this work, ~\cite{egorov2015deep} adopted DQNs to solve conventional POMDP problems.  A policy is obtained with a DQN that maps  concatenated observation-belief vector pairs to an optimal action. Their work (we call it DBQN) is designed for model-based representations of the environment where the transition, observation and reward functions are already known. Thus, the belief can be estimated precisely with Bayes' theorem and can serve as input to the neural network. However, in most real-world POMDP problems, the environment dynamics are unknown. To address this, \cite{DBLP:journals/corr/HausknechtS15}  adapted the fully connected structure of DQN with a recurrent network~\cite{DBLP:journals/neco/HochreiterS97}, and called the new architecture Deep Recurrent Q-Network (DRQN). The proposed model recurrently integrates arbitrarily long histories of observations to find an optimal policy that is robust to partial observability. However, DRQNs consider only observation histories without explicitly including actions as part of the histories. This impacts negatively the performance of the approach as demonstrated in Sec.~4. ~\cite{DBLP:journals/corr/LampleC16} combined DRQN with handcrafted features to jointly supervise the learning process of 3D games in partially observable environments, however the approach suffers from the same problem as DRQN since it overlooks action histories. ~\cite{DBLP:journals/corr/FoersterAFW16} extended DRQN to handle partially observable multi-agent reinforcement learning problems by proposing a deep distributed recurrent Q-networks (DDRQN). The action history is explicitly processed by an LSTM layer and fed as input to a Q-network. In DDRQN, each action is forcibly decoupled from its associated observation despite the fact that action-observation pairs are the key to belief updating.  As a result, the decoupling of actions and observations in DDRQN impacts negatively belief inference.

In this paper, we propose a new architecture called Action-based Deep Recurrent Q-Network (ADRQN) to improve learning performance in partially observable domains. Actions are encoded via a fully connected layer and coupled with their associated observations to form action-observation pairs. The time series of action-observation pairs is processed by an LSTM layer that learns latent states based on which a fully connected layer computes Q-values as in conventional DQNs. We demonstrate the effectiveness of our new architecture in several Atari 2600 games. Table~\ref{table:modelscomp} summarizes the main differences between ADRQN and other state-of-the-art deep Q-learning techniques.

\section{Background}
In this section, we give a brief review of Deep Q-Networks (DQNs), Partially Observable Markov Decision Processes (POMDPs) and Deep Recurrent Q-networks (DRQNs).

\subsection{Deep Q-Networks}

A sequential decision problem with known environment dynamics is usually formalized as a Markov Decision Process (MDP), which is characterized by a 4-tuple $\langle S,A,P,R \rangle$. At each step, an agent selects an action $a_t \in A$ to execute with respect to its fully observable current state $s_t \in S$ and based on its policy $\pi$.  It receives an immediate reward $r_t \sim R({s_t},{a_t})$ and transitions to a new state $s_{t+1}$. The objective of reinforcement learning is to find the policy that maximizes the expected discounted rewards $R_t$
\begin{equation}\label{equ:EDR}
{R_t} = {r_t} + \gamma {r_{t + 1}} + {\gamma ^2}{r_{t + 2}} +  \cdots
\end{equation}
where $\gamma \in [0,1]$ is the discount factor. In MDPs, an optimal policy can be computed by value iteration~\cite{DBLP:journals/tnn/SuttonB98}.

Q-Learning~\cite{DBLP:journals/ml/WatkinsD92} was proposed as a model-free technique for reinforcement learning problems with unknown dynamics.  It estimates the value of executing an action in a given state followed by an optimal policy $\pi$. This value is called the state-action value, or simply Q-value as defined below:
\begin{equation}\label{equ:QV}
{Q^\pi }(s,a) = E^{\pi}({R_t}|{s_t} = s,{a_t} = a)
\end{equation}
The Q-values can be learned iteratively according to the following rule while the agent is interacting with the environment:
\begin{equation}\label{equ:QL}
  Q(s,a) = Q(s,a) + \alpha (r + \gamma \mathop {\max }\limits_{{a'}}Q(s',a') - Q(s,a))
\end{equation}
In tasks with a large number of states, a common trick is to use a function approximator to estimate the Q-function. For instance, DQN~\cite{mnih2015human} uses a neural network parameterized by $\theta$ to represent $Q({s},{a};{\theta})$.  Neural networks with at least one (non-linear) hidden layer and sufficiently many nodes can approximate any function arbitrarily closely. DQN is optimized by minimizing the following loss function:
\begin{equation}\label{equ:DQN}
L({\theta _i}) = {E_{s,a,r,s'}}\left[ {{{\left( {(y_i^{target} - Q(s,a;{\theta _i})} \right)}^2}} \right]
\end{equation}
where $y_i^{target} = r + \gamma \mathop {\max }\limits_{{a'}} Q({s'},{a'};\theta^-_i)$ denotes the target value of the action $a_t$ given state $s_t$. Here $\theta _i^ - $ is cloned from ${\theta_i}$ every fixed numbers of iterations. DQN uses experience replay ~\cite{lin1993reinforcement} to store previous samples $e_t=\langle s_t,a_t,r_t,s_{t+1} \rangle$ up to a fixed size memory $D_t$. The Q-network is then trained by uniformly sampling mini-batches of past experiences from the replay memory. An important factor for the efficiency of DQN in AlphaGo and the Atari games is the assumption of full state observability that allows the neural network to use only one (or a few) observation(s) as input. Thus, DQN suffers inaccuracy in tasks with partially observable states.

\subsection{Partially Observable Markov Decision Processes (POMDPs)}

POMDPs generalize MDPs for planning under partial observability.  A POMDP is mathematically defined as a tuple $\langle  \mathcal{S,A,Z},T,O,R\rangle$, consisting of a finite set of states $\mathcal{S}$, a finite set of actions $\mathcal{A}$, a transition function $T : \mathcal{S}\times \mathcal{A}\rightarrow \Pi(\mathcal{S})$, where $\Pi(\mathcal{S})$ represent the set of probability distributions on $\mathcal{S}$, a reward function depending on the state and the action just performed $R
:\mathcal{S}\times \mathcal{A}\rightarrow \mathcal{R}$, a finite set of observations $\mathcal{Z}$ and an observation function $O :\mathcal{S}\times \mathcal{A}\rightarrow \Pi(\mathcal{Z})$, where $\Pi(\mathcal{Z})$ represent the set of probability distribution on $\mathcal{Z}$. As the true states are not fully observable any more, a belief is used to estimate the current state, defined as the probability mass function over the states and denoted as $b=(b(s_1),b(s_2),...b(s_{|S|}))$, where $s_i\in \mathcal{S}, b(s_i)\geq 0$, and $\sum_{s_i\in \mathcal{S}}{b(s_i)=1}$. Given the current belief $b_t$, performing action $a_t$ and get the next observation $z_{t+1}$, then the next belief $b_{t+1} = SE(b_{t},a_t,z_{t+1})$ is estimated as follows:

\begin{equation}
\nonumber b_{t+1}(s_j) =
\frac{O(s_j,a_t,z_{t+1})\sum_{s_i\in\mathcal{S}}T(s_i,a_t,s_j)b_t(s_i)}{P(z_{t+1}|a_t,b_t)}
\label{equ:SE}
\end{equation}
\begin{equation}
P(z_{t+1}|a_t,b_t)=\sum_{s_j\in\mathcal{S}}O(s_j,a_t,z_{t+1})\sum_{s_i\in\mathcal{S}}T(s_i,a_t,s_j)b_t(s_i)
\label{equ:poab} 
\end{equation}

The expected immediate reward for an agent performing action $a$ at
the belief state $b$ is computed as $\rho(b,a)=\sum_{s_i\in
\mathcal{S}}{b(s_i)R(s_i,a)}$. The transition function among beliefs
becomes:
\begin{equation}
\tau(b,a,b')=\sum_{z \in \mathcal{Z}} p(b'| b,a,z) P(z|b, a)
\label{equ:Transition} 
\end{equation}
where $p(b'| b,a,z)=1$ if $b'=SE(b,a,z)$, and $0$ otherwise.
An optimal policy $\pi:\mathcal{R}^{|\mathcal{S}|}\rightarrow
\mathcal{A}$ can be computed by value iteration:
\begin{equation} V(b) = \max\limits_a[\rho(b,a)+\gamma
\sum_{b'}{\tau(b,a,b')V(b')}] \label{eqn:pomdpvf} \vspace{-2mm}
\end{equation}
where $\gamma$ is a discount factor for the past
history. In practice, it is common to have the optimal policy
represented by a set of linear functions (so called $\alpha$-vectors) over the belief space, with the maximum ``envelop'' of the $\alpha$-vectors forming the value function~\cite{shani2013survey}.


\subsection{DRQN}
DQN assumes that the agent has full knowledge of the state of the environment, i.e., the agent's observation is equivalent to the state of environment. However, in practice, this is rarely true, even in some Atari games. For example, one frame of \emph{Pong} does not reveal the ball's velocity and its moving direction. In the game of \emph{Asteroids}, parts of the image are concealed for several consecutive frames to challenge the player. Hence, these games are naturally POMDP problems. While the strategy of DQN is to utilize several consecutive frames as the input in the hope of deducing the full state, this works well only when a short limited history is sufficient to characterize the state of the game. For more complex games, the performance of DQN decreases sharply.

To tackle problems with partially observable environments by deep reinforcement learning, ~\cite{DBLP:journals/corr/HausknechtS15} proposed a framework called \emph{Deep Recurrent Q-Learning} (DRQN) in which an LSTM layer was used to replace the first post-convolutional fully-connected layer of the conventional DQN. The recurrent structure is able to integrate an arbitrarily long history to better estimate the current state instead of utilizing a fixed-length history as in DQNs. Thus, DRQNs estimate the function $Q({o_t},{h_{t-1}}|\Theta)$ instead of $Q({s_t},{a_t}|\Theta)$, where $\Theta$ denotes the parameters of entire network, $h_{t-1}$ denotes the output of the LSTM layer at the previous step, i.e., $h_t = LSTM({h_{t-1},o_t})$. DRQN matches DQN's performance on standard MDP problems and outperforms DQN in partially observable domains. Regarding the training process, DRQN only considers the convolutional features of the history of observations instead of explicitly incorporating the actions. However as we argued in Sec.~1 and Eq.~\ref{equ:SE} in Sec.~2.2, the action performed is crucial for belief estimation. To that effect, we propose a new architecture for deep reinforcement learning with actions incorporated in histories to further improve the performance of deep RL in POMDPs. In the sequel, we will refer to our proposed architecture as ADRQN (action-specific deep recurrent Q-learning Network).

\section{ADRQN - Action-specific Deep Recurrent Q-Network}\label{ADRQN}

\begin{figure*}
  \centering
  \includegraphics[width=0.7\textwidth]{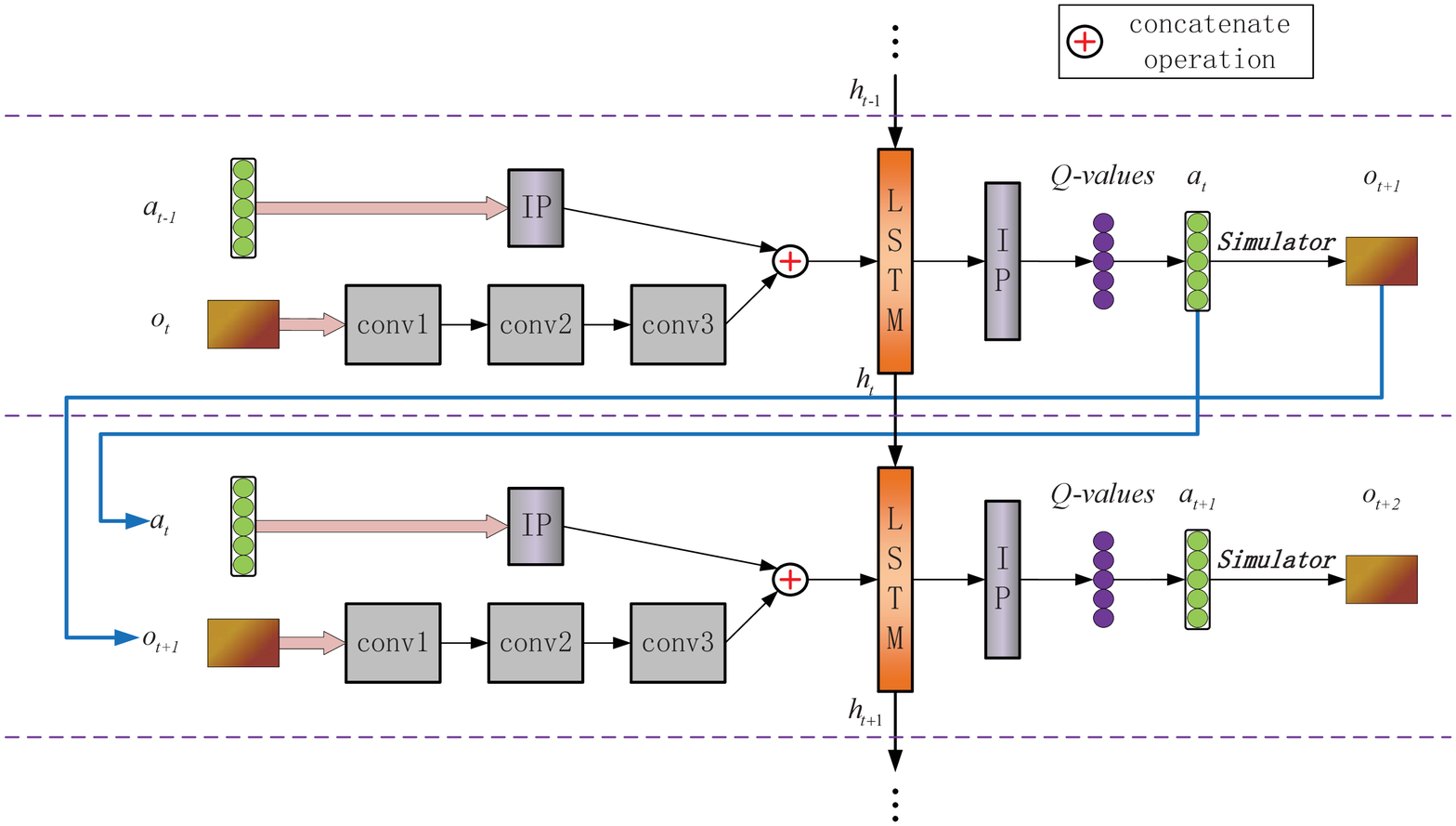}
  \caption{The input action is an 18-D vector followed by a fully connected (IP) layer with 512-D outputs. A single $84\times84$ frame will be convolved as the observation via three convolutional layers with 32 8$\times$8 filters with stride 4, 64 4$\times$4 filters with stride 2, and 64 3$\times$3 filters with stride 1, respectively. The 512-D outputs of LSTM are fed to another fully connected layer and produce 18-D Q-values corresponding to 18 actions in Atari games.}
  \vspace{0.05in}
  \label{ADRQN:architecture}
\end{figure*}

Inspired by the aforementioned works, our goal is to propose a model-free deep RL approach that incorporates the influence of the performed action through time. More specifically, we couple the performed action and the obtained observation as the input to the Q-network. The architecture\footnote{`IP' means inner product, i.e., fully connect layer}. of our model is shown in Fig.~\ref{ADRQN:architecture}.

Our first attempt to couple the action and observation was to concatenate a fixed representation of the action with the observation vector directly, i.e., we utilized one-hot vectors to represent each action $a \in \mathcal{A}$. However, such concatenation did not yield good performance because the lengths of the action and observation vectors differ largely which leads to numerical instability. In Atari games as well as many real-world POMDPs, the number of actions is far less than the dimensionality of the state representation. In conventional DQNs and DRQNs, the suggested dimensionality of the convolved features encoding the current state or observation is 3136 (after passing several convolutional and reshaping layers), while the number of actions is only 18 for the Atari games. To address this imbalance, we embed the one-hot action vectors by a fully connected layer into a higher dimensional vector (512-D is our experiments).  Thus the representations of the actions and the observations are now more balanced.

Compared with DQN and DRQN, our model is able to remember the past actions, particularly the last performed action. Thus, we modified $(s_t, a_t, r_t,s_{t+1})$ the transition in the experience replay mechanism of DQN to $\langle \{a_{t-1},o_t\},a_t,r_t,o_{t+1}\rangle$ in order to allow the framework to fetch the action-observation pair more conveniently. During the decision process for a given frame within training or the updating process of the neural network, the LSTM layer requires a sequence of action-observation pairs as its input. Thus, we store the transitions sequentially $\langle \{a_{t-1},o_t\},a_t,r_t,o_{t+1}\rangle$ within each episode in the replay memory.

Ideally, the best strategy to estimate precisely the current state for a model-free POMDP problem is to integrate the entire transition history of each episode, which usually includes thousands of transitions for an Atari game. This also means that the LSTM layer needs to be unrolled for a large number of time steps which will increase significantly the training cost. In our experiments, the LSTM layer is unrolled for 10 time steps during training. We take this setting for three points. From the perspective of learning from whole history, this setting can be more satisfy to our original idea compared to a smaller setting of the length. Second is that this setting can also guarantee the efficiency of experiments. It allows that our experiments can be finished within several days. And last point is to make a better comparison to DRQN as it take a same setting of 10. Such empirical setting has proved its efficacy to our framework.

The entire process of our proposed approach is presented in Algorithm 1. First, we initialize the parameters of the Q-network and the Target network with $\theta$ and $\theta^-$ respectively. For each episode, the first selected action is initialized to no-operation, the first hidden layer's input is initialized with a zero vector and the first observation of each episode is initialized with the preprocessed first frame. At each time step, if the observation does not indicate ``game over'' (the end of the episode), we select an action based on the $\epsilon$-greedy strategy and execute the action.  Accordingly, the immediate reward and the next observation of the screen will be obtained. The transition, once obtained, will be sequentially stored in the history of the current episode. To update the Q-network, we randomly sample a sequence of transitions $\langle a_{j-1},o_j,a_j,r_j,o_{j+1}\rangle$ to fit the unrolled LSTM layer. Then, the hidden layer $h_{j-1}$ of the Q-network and the hidden layer $h_{j}$ of the target network are obtained. The difference between these two network Q-values (i.e., Q-value $y_j$ and Q-network value $Q(h_{j-1},a_{j-1},o_j,a_j;\theta)$) is used as the loss function to update the network parameters $\theta$ via back propagation.

\renewcommand{\algorithmicrequire}{\textbf{Input:}}
\renewcommand{\algorithmicensure}{\textbf{procedure:}}
\renewcommand{\algorithmicreturn}{\textbf{Output:}}
\begin{algorithm}[t]
	\caption{Action-specific Deep Recurrent Q-Network}            	
	\label{alg:Framwork}
	\begin{algorithmic}[1]
    \State Initialize the replay memory \emph{D}, \# of iterations \emph{M}
    \State Initialize Q-Network and Target-Network with $\theta$ and $\theta^-$ respectively
    \For{episode = 1 to \emph{M}}
        \State Initialize the first action $a_0 = no\,operation$,
        ~$h_0= \textbf{0}$
        \State Init.the first obs.$o_1$ with the preprocessed first screen
        \While{$o_t \not= terminal$}
            \State Select a random action $a_t$ with the probability $\epsilon$
            \State Else select $a_t=\mathop {\argmax}_{a}Q(h_{t-1},a_{t-1},o_t,a;\theta)$
            \State Execute action $a_t$
            \State Obtain reward $r_t$ and resulting observation $o_{t+1}$
            \State Store transition $\langle \{a_{t-1},o_t\},a_t,r_t,o_{t+1}\rangle$ as one record of the current episode in \emph{D}
            \State Randomly sample a minibatch of transition
             ~~~~~~~~~~~~ sequences $\langle a_{j-1},o_j,a_j,r_j,o_{j+1}\rangle$ from \emph{D}
            \State Compute Q-value of target network
            \begin{small}
            \[y_j =\! \left\{
            {\begin{array}{*{20}l}
            r_j,&{\begin{array}{*{20}l}
            o_{j+1}\!=\!terminal
            \end{array}} \\
            r_j\!+\!\gamma \mathop {\max}\limits_{{a}}Q(h_j,a_j,o_{j\!+\!1},a;\theta^-),&{\begin{array}{*{1}l}
            o_{j+1}\!\not=\! terminal
            \end{array}} \\
            \end{array}} \right.\]
            \end{small}
            \State Compute the gradient of
           ~~~~~~~~~~~~~~~$(y_j-Q(h_{j-1},a_{j-1},o_{j},a_j;\theta))^2$ to update $\theta$
        \EndWhile
    \EndFor
	\end{algorithmic}
\end{algorithm}

\section{Experiments}

We evaluate the training and testing performance of ADRQN with several Atari games and their flickering version.

\subsection{Experiments setup}
\begin{itemize}
  \item \textbf{Flickering Atari game:} ~\cite{DBLP:journals/corr/HausknechtS15} introduced a flickering version of the Atari games, which modified the Atari games by obscuring the entire screen with a certain probability at each time step, which introduces partial observability and therefore yields a POMDP. In their framework, before a frame is sent to the neural network as input, each raw screen is either fully observable or fully obscured with black pixels.

  \item \textbf{Frame skip Scheme}. We adopted the frame skip technique~\cite{DBLP:journals/jair/BellemareNVB13}. This mechanism is commonly used in most previous works of deep reinforcement learning to efficiently simulate the environment. In this mechanism, an agent performs the selected action $a_i$ for $k+1$ consecutive frames and treats the transition from the first frame $f_0$ to frame $f_{k+1}$ as the effect of action $a_i$, i.e. $\langle f_0,a_i, f_1,...,a_i,f_{k+1} \rangle$=$\langle f_0, a_i, f_{k+1} \rangle$. Thus, a large number of frames can be skipped to accelerate the training process, but with a tolerable performance loss. In our experiments, $k$ is set to 4.
  \item \textbf{Hyperparameters}. In our experiments, we also adopt experience replay and set the replay memory size to $\emph{D}=400,000$ (i.e., storing 400,000 transitions). When selecting an action, we follow the $\epsilon\!-\!greedy$ policy with $\epsilon=1- \frac{0.9*iter}{explore}$, where $iter$ is the current number of iterations performed and $explore$ is the number of iterations that epsilon reaches to a given value. In our setting,  $\epsilon$ will reach 0.1 and $explore$ was set to $1,000,000$. The discount factor $\gamma$ was set to 0.99. The target network is updated by cloning the weights of the Q-network every $10,000$ iterations. And we unrolled the LSTM to 10 time step when training as we said in Section~\ref{ADRQN}.

  \item \textbf{Random Updates}. As suggested in~\cite{DBLP:journals/corr/HausknechtS15}, random updates can achieve the similar performance as conventional sequential updates of the entire episode, but with much lower training cost. Random updates consist of selecting randomly a series of transitions from one episode as the input. In our framework, this corresponds to utilizing a sequence of action-observation pairs to perform the updates. The initial hidden input $h_0$ of the RNN is set to the zero vector at the beginning of the update.

\end{itemize}

\subsection{Training}

\begin{figure}[h]
\setlength{\abovecaptionskip}{0.cm}
  \centering
  \includegraphics[width=0.37\textwidth]{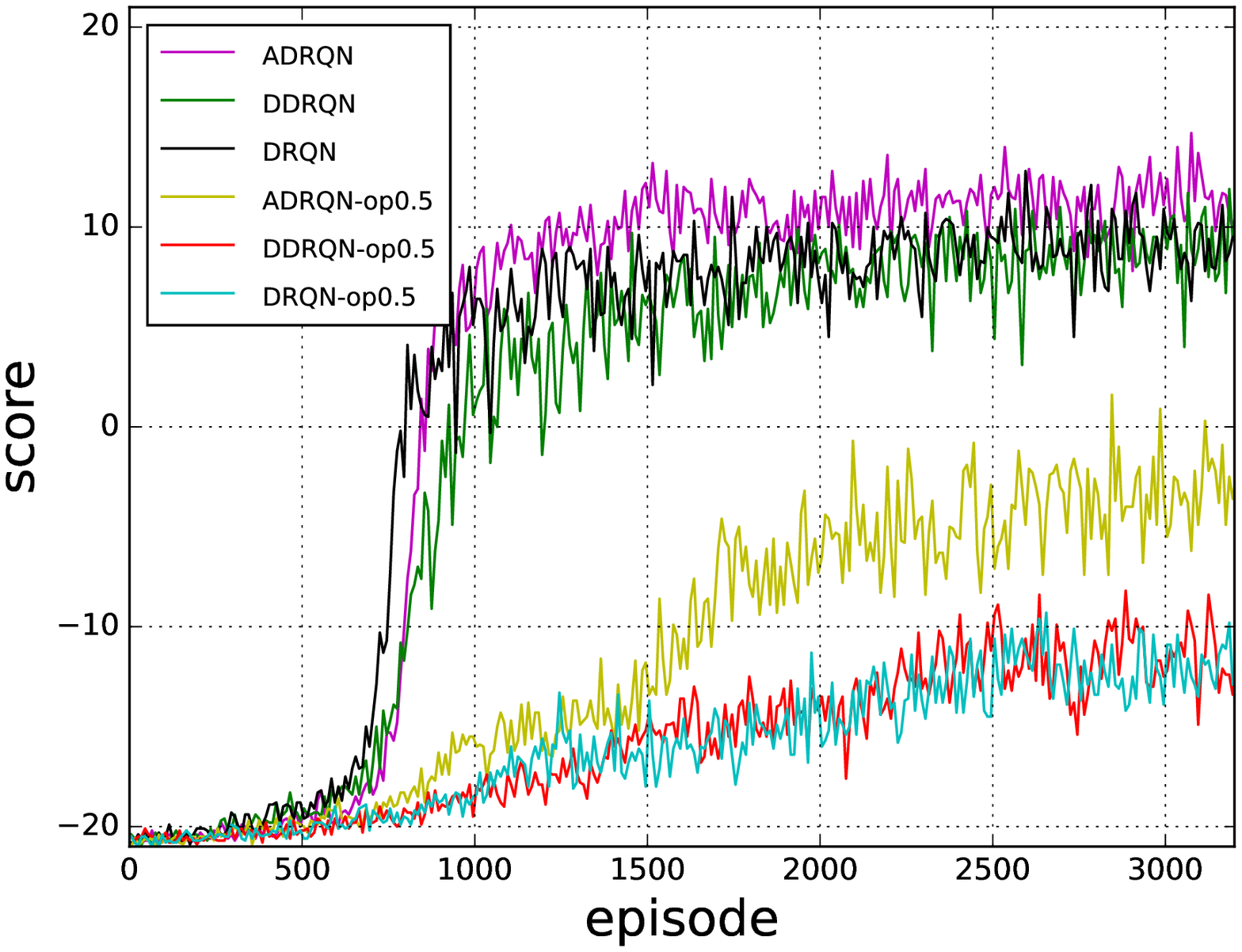}
  \caption{Training results for Pong}
  \label{pong:training}

\setlength{\abovecaptionskip}{0.cm}
  \centering
  \includegraphics[width=0.37\textwidth]{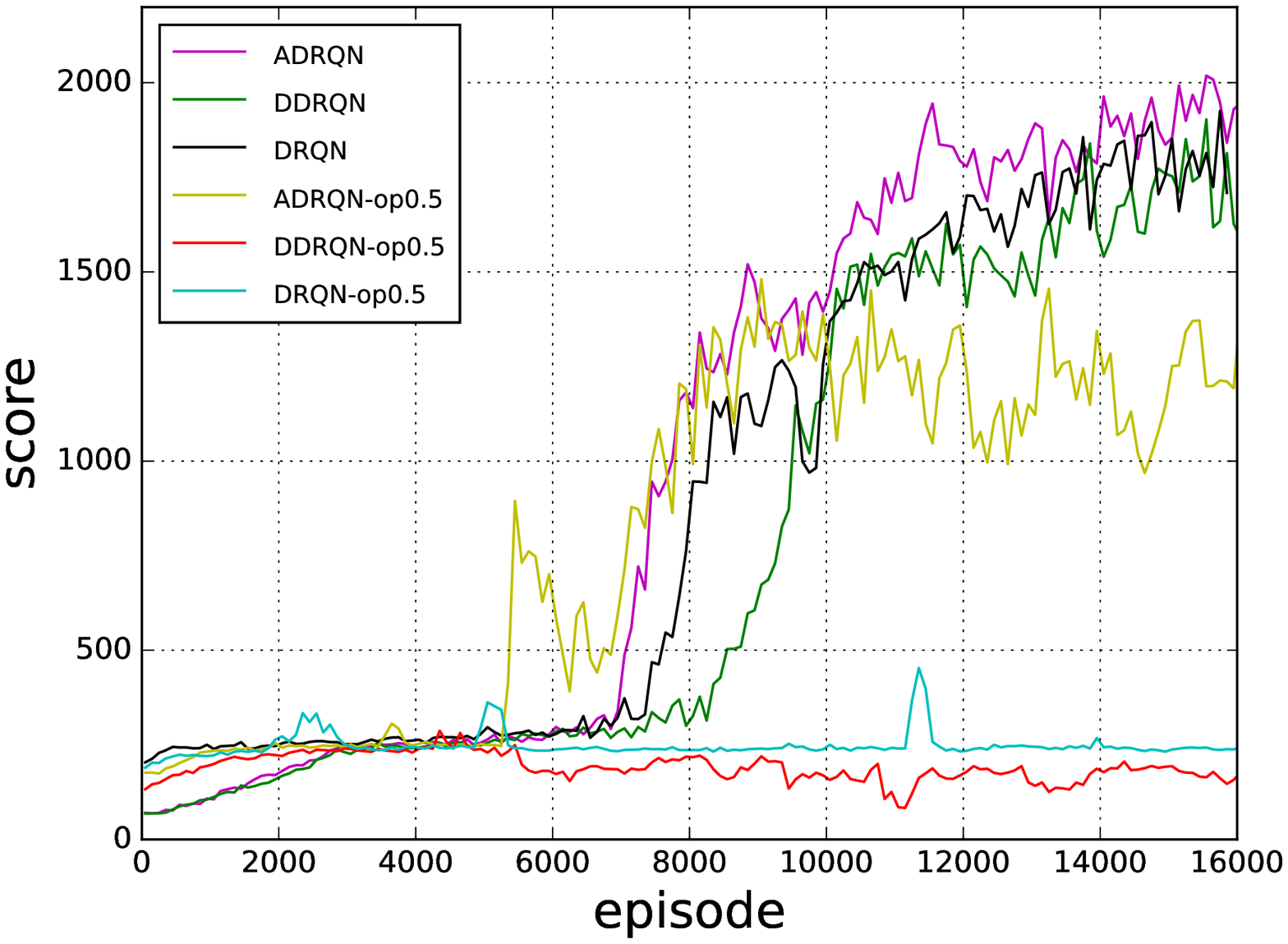}
  \caption{Training results for Frostbite}
  \label{Frostbite:training}
\end{figure}
\vspace{-1mm}
Compared to the other approaches, the key idea of our framework is to construct action-observation pairs as input to the LSTM layer to retain more representative features for the Q-network to learn recurrently. The actions are first encoded with one-hot vectors,
then processed by a fully connected layer to construct a higher-dimensional vector that is concatenated with the output of the third convolutional layer for better numerical stability. As the LSTM layer will be unrolled for 10 time steps, we need to ensure that there are enough transitions to be stored in the replay memory \emph{D} so that we can sample a minibatch of the transition sequences of length 10 each time we update the entire network. In our experiments, we update the entire network until the replay memory is full. Moreover, the scores obtained by playing the games are not always stable since small changes of the weights may have a significant impact on the outcome ~\cite{DBLP:journals/corr/MnihKSGAWR13}. And, it may bring the instability to the policy which the network have learned.
Thus, we adopt an adaptive setting as done in most previous works by replacing negative rewards by -1 and positive rewards by +1.

When training the flickering versions of Atari games, we set the probability of obscuring a frame to 0.5 as a compromise. A lower probability may prevent learning due to a large loss of information, and a higher probability may be less convincing that the transformed version is a POMDP. Besides, it is a fair setting to evaluate generalization performance since it divides the test interval into two subintervals evenly. The games \emph{Pong} and \emph{Frostbite} are both trained under the setting of full observation and a 0.5 probability of obscuring a frame. The training performances of \emph{Pong} and \emph{Frostbite} are shown in Fig.~\ref{pong:training} and~\ref{Frostbite:training} respectively. The reported scores are based on averages of 10 and 100 episodes respectively.

We compared our model ADRQN with DRQN and DDRQN on several Atari games. As DDRQN is proposed for addressing multi-agent POMDP problems and it disabled experience replay mechanism, we adapted it to a single-agent version which also adopts the replay memory mechanism. Fig.~\ref{pong:training} and Fig.~\ref{Frostbite:training} show that ADRQN matches the performance obtained by DRQN and DDRQN, and even performs slightly better than them in the conventional setting (full observation), while the improvements of ADRQN over DRQN and DDRQN in partially observable settings are very obvious. 
Additionally, when trained under the setting of the observation probability as 0.5, ADRQN can obtain significant improvements in \emph{Frostbite}(Fig.~\ref{Frostbite:training}). We believe the gain comes from the use of $\langle$action, observation$\rangle$ pair which speeds up the training process in partially observable environments. In general, our proposed model can lead to higher scores in partially observable settings during the training process, which supports the argument that action-observation pairs are important in POMDP problems.


\subsection{Testing Evaluation}
\begin{table}
\begin{scriptsize}
\center
    \begin{tabular}{|c|c|c|c|}
        \hline
            Model & DRQN($\pm$\emph{std}) & DDRQN($\pm$\emph{std}) & ADRQN($\pm$\emph{std})\\
            \hline
             Pong & $18.3(\pm2.0)$ & $18.6(\pm1.2)$ & $18.54(\pm2.3)$\\
            \hline
            Chp.Cmd. & $1790(\pm744.3)$ & $1455(\pm596.0)$ & $1648(\pm658.1)$\\
            \hline
           Asteroids & $983.8(\pm366.9)$ & $1096.6(\pm351.9)$ &$1025.4(\pm360.9)$\\
           \hline
           Double dunk & $-12.8(\pm3.8)$ & $-13(\pm4.5)$ & $-15.2(\pm3.4)$\\
           \hline
           Frostbite & $2412(\pm394.5)$ & $2245.5(\pm585.8)$ & $2290.5(\pm571.7)$\\
        \hline
    \end{tabular}
    \caption{Testing results from standard setting}
    \label{evaluation-op1}
\end{scriptsize}
\end{table}

\begin{table}
\begin{scriptsize}
\center
    \begin{tabular}{|c|c|c|c|}
        \hline
            Model & DRQN($\pm$\emph{std}) & DDRQN($\pm$\emph{std}) & ADRQN($\pm$\emph{std})\\
            \hline
             Pong & $1.6(\pm7.8)$ & $1.9(\pm8.4)$ & $\textbf{7}(\pm4.6)$\\
            \hline
            Chp. Cmd. & $1090(\pm409.2)$ & $1040(\pm392.8)$ & $\textbf{1608}(\pm707.9)$\\
            \hline
           Asteroids & $871.4(\pm339.8)$ & $1033(\pm396.1)$ &$\textbf{1040.2}(\pm431.5)$\\
           \hline
           Double dunk & $-14.4(\pm3.2)$ & $-13(\pm2.5)$ & $-13(\pm3.6)$\\
           \hline
           Frostbite & $673.5(\pm503.0)$ & $393(\pm347.4)$ & $\textbf{2002.5}(\pm734.653)$\\
        \hline
    \end{tabular}
    \caption{Testing results with flickering setting (obs. prob. is 0.5)}
    \label{evaluation-op0.5}
\end{scriptsize}
\end{table}
We also evaluate the well-learned DRQN, DDRQN and ADRQN models on the five games and their flickering version. We replay each game based on the learned model 50 times to obtain the average scores as our final results. The epsilon value used in testing evaluation for $\epsilon\!-\!greedy$ is set to 0, as we consider the model has been well trained and there is no need for the exploration. Table~\ref{evaluation-op1} summarizes the results obtained with full observations. The results shows that ADRQN has a similar performance with DRQN and DDRQN in general in full observable environments. While Table~\ref{evaluation-op0.5} demonstrates that ADRQN significantly outperforms DRQN and DDRQN when trained with half of the frames obscured in all five games. And particularly, that ADRQN can get a similar score as it get in full observable setting in the game Chopper Command and game Frostbite which further demonstrates the advantage of our model.
An interesting observation is that the testing results are generally better than the results obtained during the training process for all three models. This may be explained by the fact that different distribution of the concealed frames may render the problem more or less observable, to the extent that better results may be obtained in some cases. Another explanation is that no exploration is needed for selecting action or making a decision after a policy is well trained, thus leads to better results.

\subsection{Generalization Performance}
To further demonstrate the advantages of ADRQN in dealing with environmental changes, we evaluate the generalization performance of DRQN, DDRQN and ADRQN on the flicking versions of the games.

\begin{figure}[h]
	\centering
	\subfigure[Game Pong]{\includegraphics[width=1.6in]{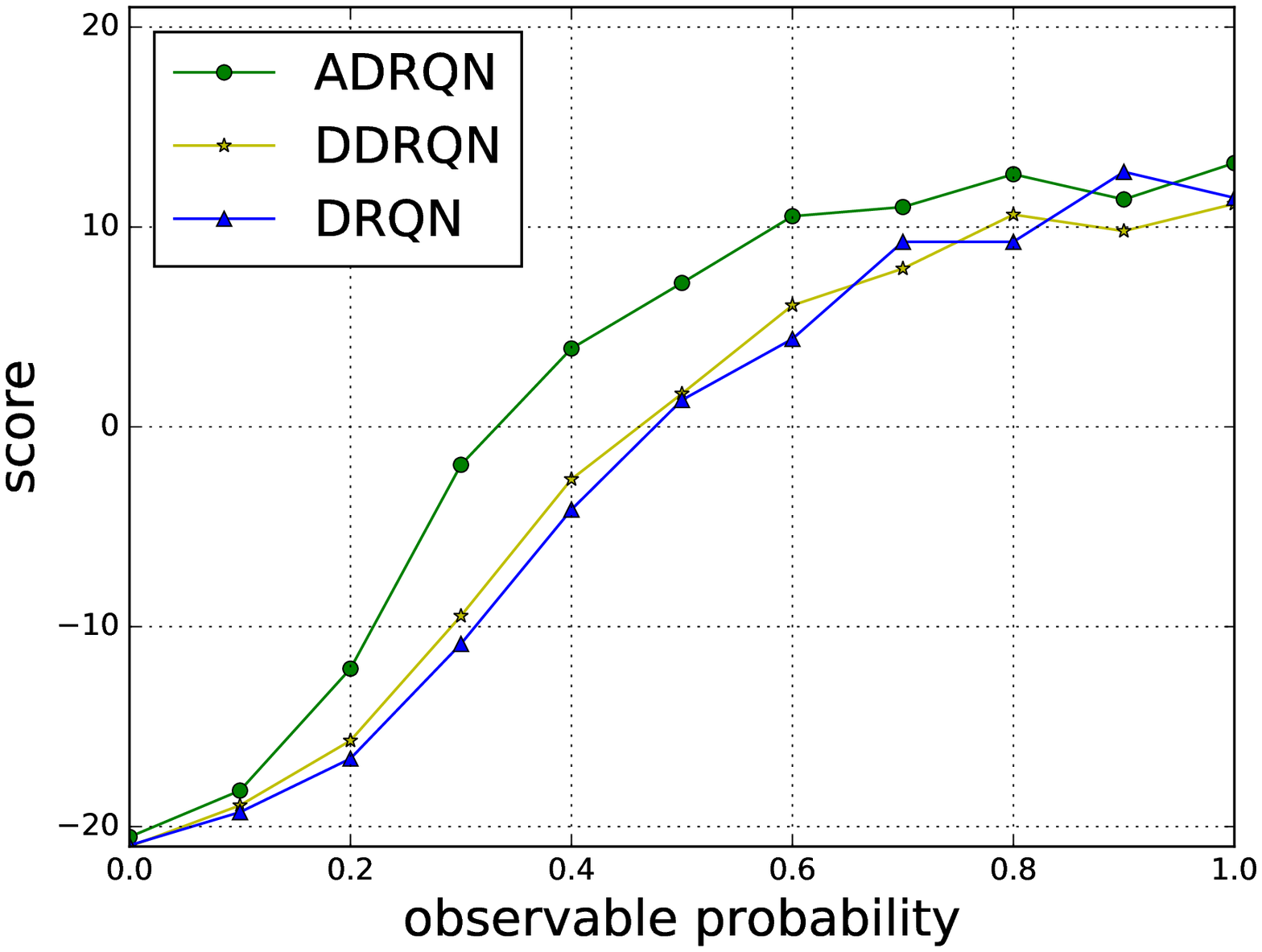}\label{pong:generalization} }
	\subfigure[Game Frostbite]{\includegraphics[width=1.6in]{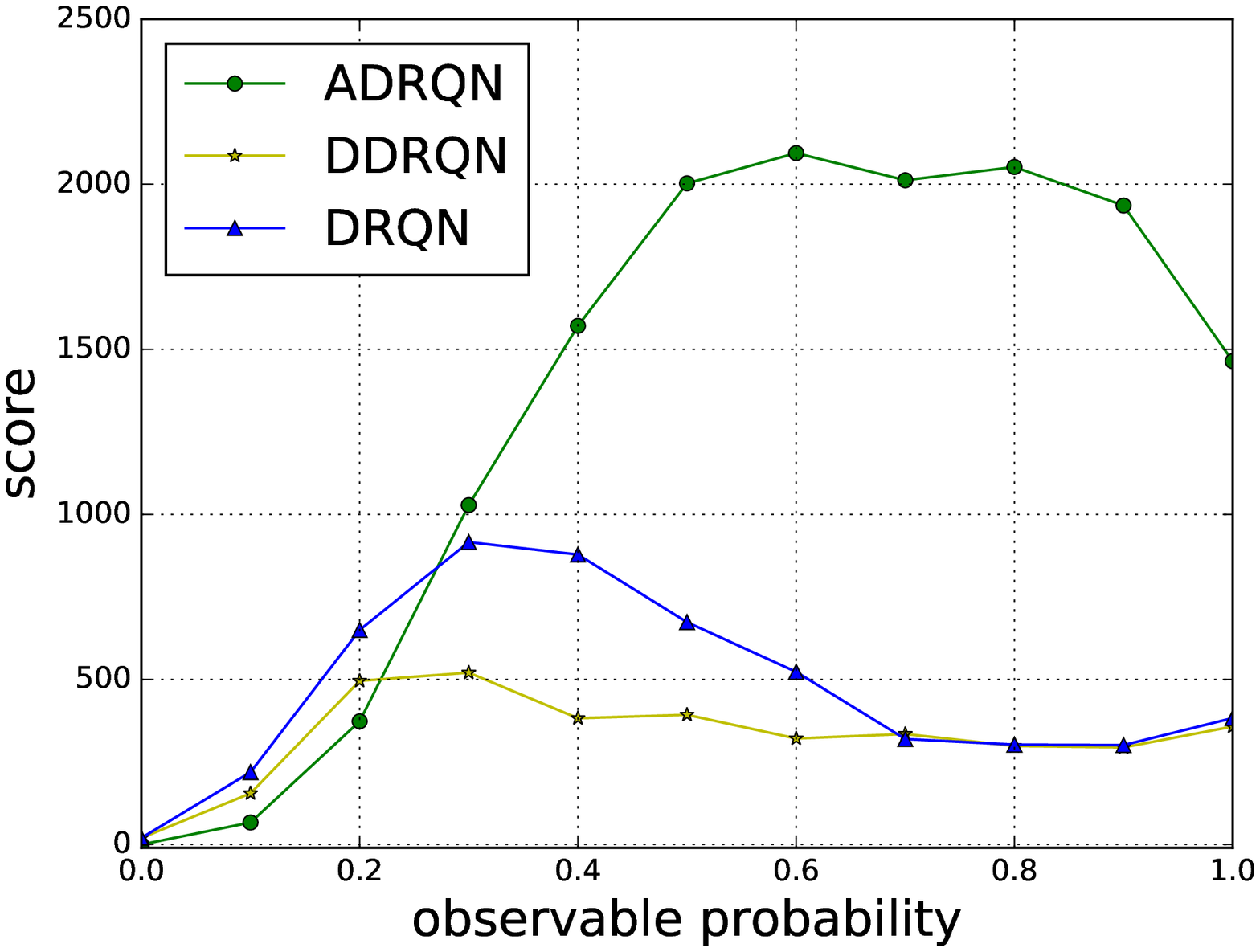}\label{Frostbite:generalization} }
	\caption{POMDP to MDP Generalization Performance}\label{fig:generalization}

	\centering
	\subfigure[Game Pong]{\includegraphics[width=1.6in]{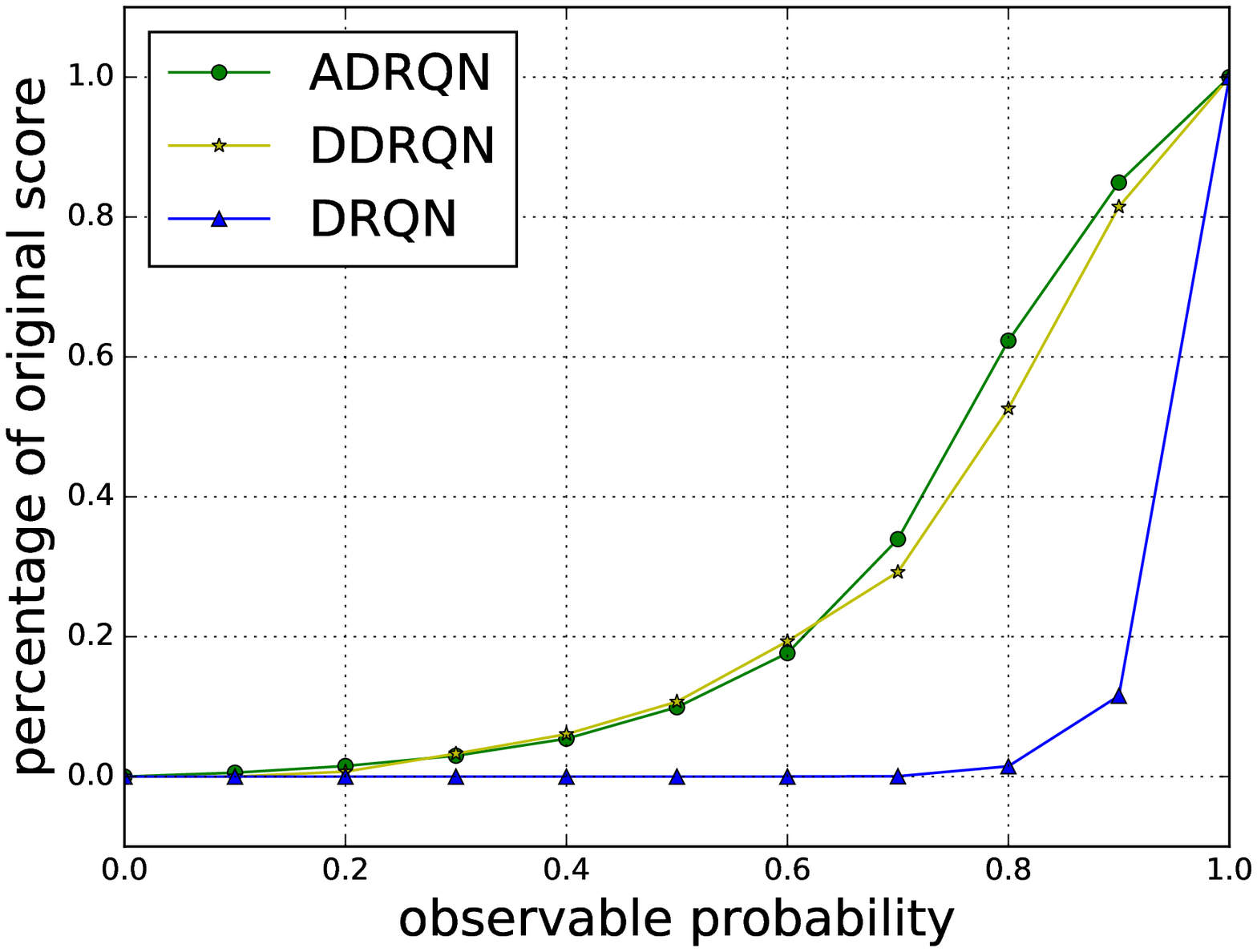}\label{pong:robustness} }
	\subfigure[Game Frostbite]{\includegraphics[width=1.6in]{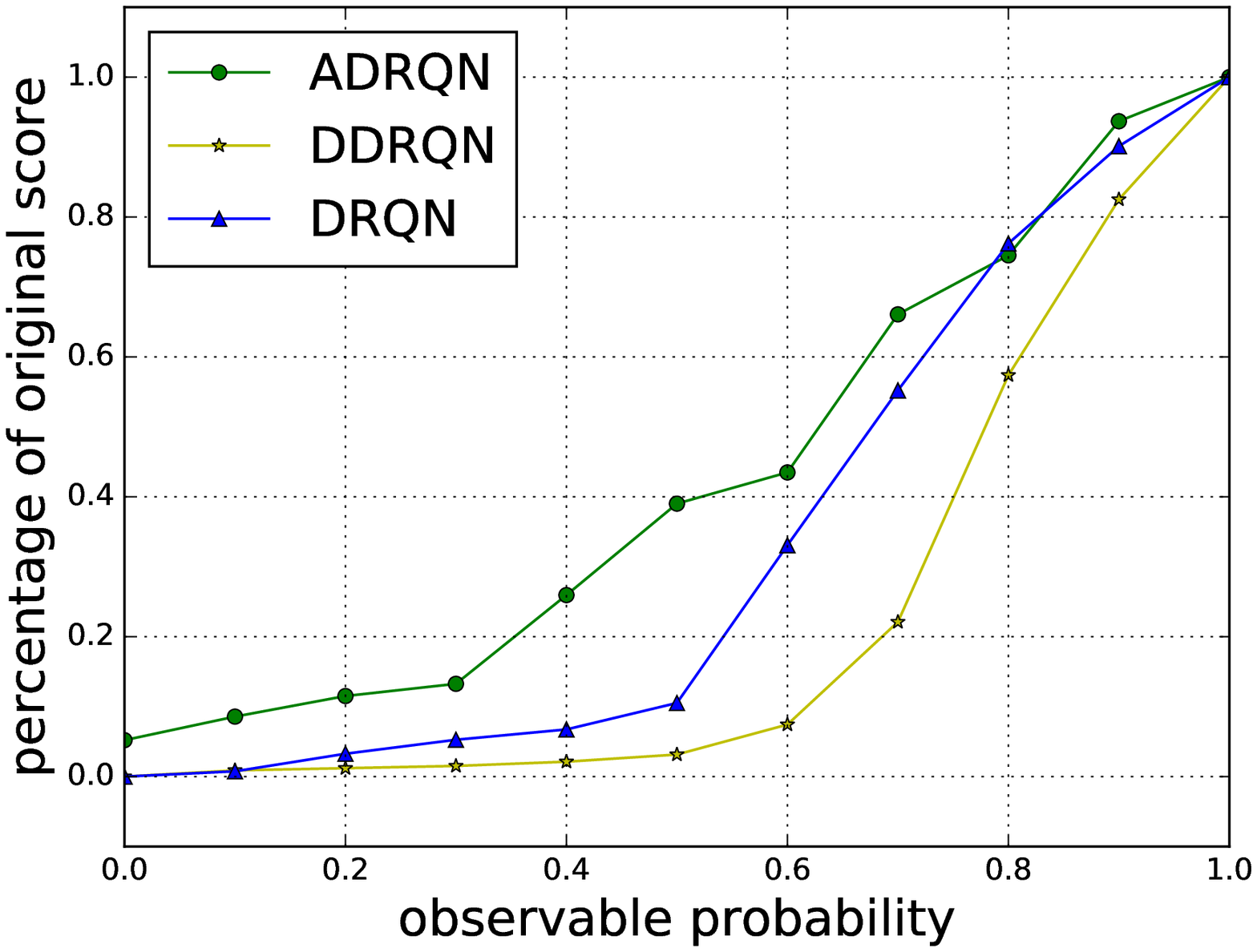}\label{Frostbite:robustness} }
	\caption{MDP to POMDP Generalization Performance}\label{fig:Robustness}
\end{figure}

\textbf{POMDP to MDP Generalization:} After being trained with observation probability of 0.5, we test the learned policy in settings with observation probabilities $(0, 0.1, 0.2, 0.3, 0.4, 0.5, 0.6, 0.7, 0.8, 0.9, 1.0)$. The results are shown in Fig.~\ref{fig:generalization}. We observe that all of three models consistently perform better with the increase of observation probability in game Pong, and also perform better with the observation probability changed in game Frostbite.
ADRQN consistently outperforms DRQN and DDRQN in these two games, and has a large advantage in game Frostbite which further demonstrates that our model can better respond to environmental changes. 

\textbf{MDP to POMDP Generalization:} After being trained as an MDP by setting the observation probability to 1, we test the learned policy with different settings of the obscured probability. As shown in Fig.\ref{fig:Robustness}, all the performances achieved by the three models drop sharply as we increase the amount of missing information.  Fig.\ref{fig:Robustness} clearly demonstrate that ADRQN is relatively robust compared to DRQN and DDRQN.

\vspace{-5mm}
\section{Brief Discussion}
Someone may argue that action information could be viewed as redundant in the context of a fixed policy. That is if you know the entire sequence of observations and there is a fixed deterministic policy, then one can fill in the missing actions directly. We agree with this argument and we guess that's why the existing state-of-the-art work i.e. DRQN overlooked the action observation pair. However, with the adoption of a recurrent model, the observation history is inevitably truncated to ensure training efficiency. Thus, a proper action sequence cannot be inferred from the truncated history. For a complex partially observable task, the set of observations is infinite. It would be even harder to infer the actions from the finite observation history. Intuitively, an explicit incorporation of the coupled action-observation pair would help a lot especially during the training process.

\section{Conclusion}

In this paper, we propose an action-specific deep recurrent Q-Network (ADRQN) to enhance the learning performance in partially observable domains. We first encode actions via MLP and couple them with the features of observations extracted from CNN to form action-observation pairs. Then an LSTM layer is utilized to integrate a time series of action-observation pairs to infer the latent states. Finally, a fully connected layer will address Q-values computing to guide the overall learning process which is similar to conventional DQNs. We have demonstrated the effectiveness of our proposed approach in several POMDP problems in comparison to the state-of-the-art approaches.



\bibliographystyle{named}
\bibliography{ijcai18}

\begin{thebibliography}{}

\bibitem[\protect\citeauthoryear{Bellemare \bgroup \em et al.\egroup
  }{2013}]{DBLP:journals/jair/BellemareNVB13}
Marc~G. Bellemare, Yavar Naddaf, Joel Veness, and Michael Bowling.
\newblock The arcade learning environment: An evaluation platform for general
  agents.
\newblock {\em J. Artif. Intell. Res. {(JAIR)}}, 47:253--279, 2013.

\bibitem[\protect\citeauthoryear{Egorov}{2015}]{egorov2015deep}
Maxim Egorov.
\newblock Deep reinforcement learning with pomdps.
\newblock 2015.

\bibitem[\protect\citeauthoryear{Foerster \bgroup \em et al.\egroup
  }{2016}]{DBLP:journals/corr/FoersterAFW16}
Jakob~N. Foerster, Yannis~M. Assael, Nando de~Freitas, and Shimon Whiteson.
\newblock Learning to communicate to solve riddles with deep distributed
  recurrent q-networks.
\newblock {\em CoRR}, abs/1602.02672, 2016.

\bibitem[\protect\citeauthoryear{Hausknecht and
  Stone}{2015}]{DBLP:journals/corr/HausknechtS15}
Matthew~J. Hausknecht and Peter Stone.
\newblock Deep recurrent q-learning for partially observable mdps.
\newblock {\em CoRR}, abs/1507.06527, 2015.

\bibitem[\protect\citeauthoryear{Hochreiter and
  Schmidhuber}{1997}]{DBLP:journals/neco/HochreiterS97}
Sepp Hochreiter and J{\"{u}}rgen Schmidhuber.
\newblock Long short-term memory.
\newblock {\em Neural Computation}, 9(8):1735--1780, 1997.

\bibitem[\protect\citeauthoryear{Kurniawati \bgroup \em et al.\egroup
  }{2008}]{DBLP:conf/rss/KurniawatiHL08}
Hanna Kurniawati, David Hsu, and Wee~Sun Lee.
\newblock {SARSOP:} efficient point-based {POMDP} planning by approximating
  optimally reachable belief spaces.
\newblock In {\em Robotics: Science and Systems IV, Eidgen{\"{o}}ssische
  Technische Hochschule Z{\"{u}}rich, Zurich, Switzerland, June 25-28, 2008},
  2008.

\bibitem[\protect\citeauthoryear{Lample and
  Chaplot}{2016}]{DBLP:journals/corr/LampleC16}
Guillaume Lample and Devendra~Singh Chaplot.
\newblock Playing {FPS} games with deep reinforcement learning.
\newblock {\em CoRR}, abs/1609.05521, 2016.

\bibitem[\protect\citeauthoryear{Li \bgroup \em et al.\egroup
  }{2007}]{DBLP:conf/icml/LiCLW07}
Xin Li, William~Kwok{-}Wai Cheung, Jiming Liu, and Zhili Wu.
\newblock A novel orthogonal nmf-based belief compression for pomdps.
\newblock In {\em Machine Learning, Proceedings of the Twenty-Fourth
  International Conference {(ICML} 2007), Corvallis, Oregon, USA, June 20-24,
  2007}, pages 537--544, 2007.

\bibitem[\protect\citeauthoryear{Lillicrap \bgroup \em et al.\egroup
  }{2015}]{DBLP:journals/corr/LillicrapHPHETS15}
Timothy~P. Lillicrap, Jonathan~J. Hunt, Alexander Pritzel, Nicolas Heess, Tom
  Erez, Yuval Tassa, David Silver, and Daan Wierstra.
\newblock Continuous control with deep reinforcement learning.
\newblock {\em CoRR}, abs/1509.02971, 2015.

\bibitem[\protect\citeauthoryear{Lin}{1993}]{lin1993reinforcement}
Long-Ji Lin.
\newblock Reinforcement learning for robots using neural networks.
\newblock Technical report, DTIC Document, 1993.

\bibitem[\protect\citeauthoryear{Mnih \bgroup \em et al.\egroup
  }{2013}]{DBLP:journals/corr/MnihKSGAWR13}
Volodymyr Mnih, Koray Kavukcuoglu, David Silver, Alex Graves, Ioannis
  Antonoglou, Daan Wierstra, and Martin~A. Riedmiller.
\newblock Playing atari with deep reinforcement learning.
\newblock {\em CoRR}, abs/1312.5602, 2013.

\bibitem[\protect\citeauthoryear{Mnih \bgroup \em et al.\egroup
  }{2015}]{mnih2015human}
Volodymyr Mnih, Kavukcuoglu Koray, Silver David, Rusu~Andrei A, Veness Joel,
  Bellemare~Marc G, Graves Alex, Riedmiller Martin, Fidjeland~Andreas K,
  Ostrovski Georg, et~al.
\newblock Human-level control through deep reinforcement learning.
\newblock {\em Nature}, 518(7540):529--533, 2015.

\bibitem[\protect\citeauthoryear{Shani \bgroup \em et al.\egroup
  }{2013}]{shani2013survey}
Guy Shani, Joelle Pineau, and Robert Kaplow.
\newblock A survey of point-based pomdp solvers.
\newblock {\em Autonomous Agents and Multi-Agent Systems}, pages 1--51, 2013.

\bibitem[\protect\citeauthoryear{Silver \bgroup \em et al.\egroup
  }{2016}]{silver2016mastering}
David Silver, Aja Huang, Chris~J Maddison, Arthur Guez, Laurent Sifre, George
  Van Den~Driessche, Julian Schrittwieser, Ioannis Antonoglou, Veda
  Panneershelvam, Marc Lanctot, et~al.
\newblock Mastering the game of go with deep neural networks and tree search.
\newblock {\em Nature}, 529(7587):484--489, 2016.

\bibitem[\protect\citeauthoryear{Stadie \bgroup \em et al.\egroup
  }{2015}]{DBLP:journals/corr/StadieLA15}
Bradly~C. Stadie, Sergey Levine, and Pieter Abbeel.
\newblock Incentivizing exploration in reinforcement learning with deep
  predictive models.
\newblock {\em CoRR}, abs/1507.00814, 2015.

\bibitem[\protect\citeauthoryear{Sutton and
  Barto}{1998}]{DBLP:journals/tnn/SuttonB98}
Richard~S. Sutton and Andrew~G. Barto.
\newblock Reinforcement learning: An introduction.
\newblock {\em {IEEE} Trans. Neural Networks}, 9(5):1054--1054, 1998.

\bibitem[\protect\citeauthoryear{Watkins and
  Dayan}{1992}]{DBLP:journals/ml/WatkinsD92}
Christopher J. C.~H. Watkins and Peter Dayan.
\newblock Technical note q-learning.
\newblock {\em Machine Learning}, 8:279--292, 1992.

\end{thebibliography}

\end{document}